\documentclass{article} %

\PassOptionsToPackage{table}{xcolor}

\usepackage[preprint]{colm2026_conference}

\usepackage{microtype}
\usepackage{graphicx}
\usepackage{subcaption}
\usepackage{booktabs}
\usepackage{amsmath}
\usepackage{amssymb}
\usepackage{mathtools}
\usepackage{amsthm}
\usepackage{hyperref}
\usepackage{url}
\usepackage{lipsum}
\usepackage{multirow}
\usepackage[normalem]{ulem}
\usepackage{soul}
\usepackage{tcolorbox}
\usepackage{wrapfig}
\usepackage{circledsteps}
\usepackage{xcolor}
\usepackage{colortbl}
\usepackage[utf8]{inputenc}
\usepackage[T1]{fontenc}
\usepackage{nicefrac}
\usepackage[capitalize,noabbrev]{cleveref}
\usepackage[textsize=tiny]{todonotes}
\usepackage{lineno}

\theoremstyle{plain}
\theoremstyle{definition}
\theoremstyle{remark}

\definecolor{mintbg}{rgb}{.63,.79,.95}
\colorlet{lightmintbg}{mintbg!50}
\colorlet{lightlightgray}{lightgray!40}
\colorlet{lightyellow}{yellow!40}

\definecolor{darkblue}{rgb}{0, 0, 0.5}
\hypersetup{colorlinks=true, citecolor=darkblue, linkcolor=darkblue, urlcolor=darkblue}

\renewcommand{\comment}[1]{}

\title{TokenButler: Token Importance is Predictable}

\author{Yash Akhauri$^{1*}$, Ahmed F. AbouElhamayed$^{1*}$, Yifei Gao$^{1}$, Chi-Chih Chang$^{1}$, \\
\bfseries Sameh Gobriel$^{2}$, Nilesh Jain$^{2}$, Mohamed S. Abdelfattah$^{1}$ \\
$^{1}$Cornell University \quad $^{2}$Intel Labs
}

\begin{document}

\ifcolmsubmission
\linenumbers
\fi

\maketitle
{\renewcommand{\thefootnote}{*}\footnotetext{Equal contribution.}}

\begin{abstract}
Large Language Models (LLMs) rely on the Key-Value (KV) Cache to store token history, enabling efficient decoding of tokens.
As the KV-Cache grows, it becomes a major memory and computation bottleneck. However, there is an opportunity to alleviate this bottleneck, prior research has shown that only a small subset of tokens contribute meaningfully to each decoding step.
A key challenge in finding these \textit{critical tokens} is that they are dynamic, and heavily input query-dependent.
Existing methods either risk quality by evicting tokens permanently, or retain the full KV-Cache but rely on retrieving chunks (pages) of tokens at generation, failing at dense, context-rich tasks.
Additionally, many existing KV-Cache sparsity methods rely on inaccurate proxies for token importance.
To address these limitations, we introduce \textbf{TokenButler}, a high-granularity, query-aware predictor that learns to identify these critical tokens.
TokenButler predicts low-dimensional \emph{importance queries} at a fixed depth stride, and combines them with a learned projection of the \emph{real KV-cache keys} to score tokens cheaply, enabling dynamic per-token selection under a fixed budget while preserving the full KV cache. We train TokenButler by distilling the model's masked causal attention distributions, optimizing a lightweight predictor with minimal parameter overhead.
We evaluate TokenButler on a novel synthetic small-context co-referential retrieval task, demonstrating near-oracle accuracy where existing methods fail. Furthermore, TokenButler achieves competitive or superior performance on long-context benchmarks (RULER, LongBench), up to $\approx1.6\times$ on-GPU speedup using our proposed \emph{prediction interval with neighbor fetching} that amortizes predictor cost while maintaining accuracy within $\approx$1.1\%, and up to 7.6$\times$ reduction in latency compared to Dense Attention with CPU offloading, enabling efficient decoding for long contexts in limited GPU memory systems. Code \href{https://github.com/abdelfattah-lab/TokenButler}{\texttt{is available}}.
\end{abstract}

\section{Introduction}

As Large Language Models (LLMs) become more widely used~\citep{thoppilan2022lamda,yuan2022wordcraft,wei2022emergent,zhang2023benchmarking}, recent advances have extended their context lengths to 64k–1M tokens. The Key-Value (KV) cache grows linearly with the context length, becoming a major memory and bandwidth bottleneck. Prior work addresses this via sparsity, quantization, efficient attention, or low-rank compression~\citep{child2019generating,choromanski2020rethinking,katharopoulos2020transformers,shazeer2019fast,pope2022efficiently,sun2024shadowkvkvcacheshadows,akhauri2024attambaattendingmultitokenstates,chen2025powernegativezerodatatype}.

Token pruning methods fall into three categories:
\textbf{(1)}~\textit{Static strategies} that cap the KV-Cache with fixed rules on removing tokens (StreamingLLM~\citep{xiaoefficientstreamllm}, Sliding Window~\citep{luong2015effective_swa});
\textbf{(2)}~\textit{Adaptive eviction} that permanently drops low-importance tokens ($H_{2}O$, SnapKV~\citep{zhang2023h2o, li2024snapkv}); and
\textbf{(3)}~\textit{Adaptive dynamic strategies} that preserve the full cache but selectively access a subset at decode time, reducing bandwidth at the cost of higher storage (Quest~\citep{tang2024quest}).

\begin{figure*}[t]
    \centering
    \includegraphics[width=\linewidth]{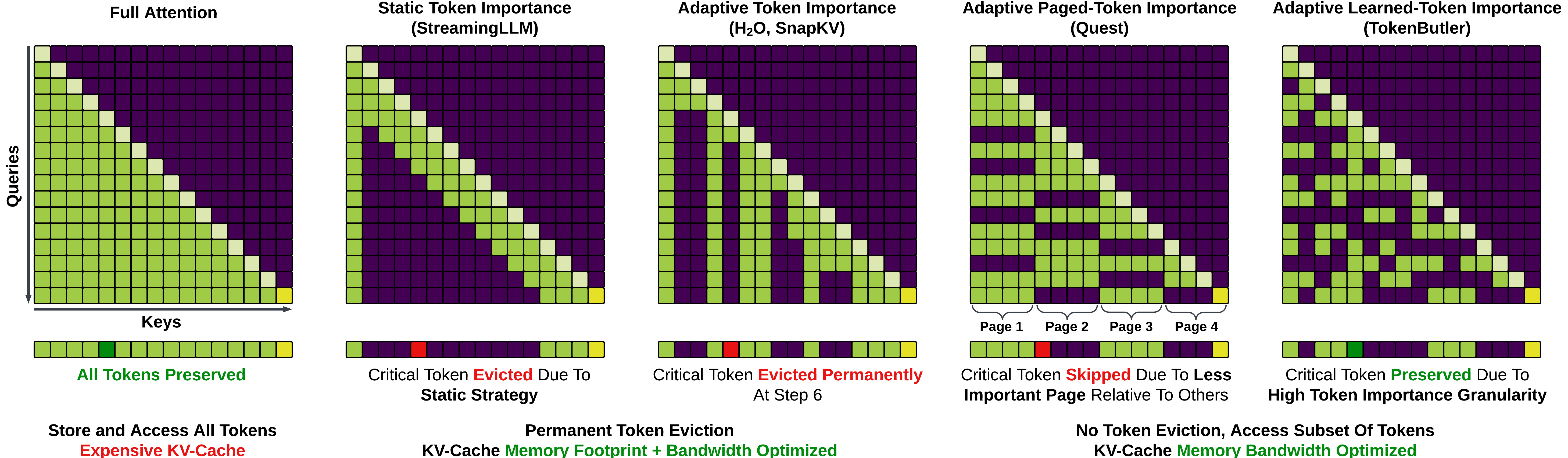}
    \caption{Full-Attention preserves all tokens, enabling access to the critical token (dark green) during the last decode step. Static strategies like StreamingLLM will not be able to access this token. Methods like $H_{2}O$ may have evicted the token at an earlier decode step, if deemed unimportant. Paged-Token importance may cause a page-miss of a critical token in context dense tasks. \textbf{TokenButler} can effectively predict critical tokens, and can be leveraged by existing methods to offer both high-granularity and cheap importance estimation.}
    \label{fig:TokenImpDemo_v2}
\end{figure*}

Token importance is highly query-dependent~\citep{tang2024quest} and each strategy has its limitations: static methods lack query-awareness, and adaptive eviction permanently discards tokens that may become relevant again in \textit{co-referential} contexts~\citep{vodrahalli2024michelangelolongcontextevaluations}. A context is co-referential when text introduced earlier is referenced again later, requiring accurate retrieval over the earlier mention. In such scenarios, token importance is non-monotonic: a critical entity may remain dormant for thousands of steps, only to become the most relevant token when a later query triggers a callback. Adaptive dynamic strategies address this by preserving the full cache, but current methods rely on coarse \textit{token grouping} for efficiency~\citep{tang2024quest}. Figure~\ref{fig:TokenImpDemo_v2} illustrates these failure modes.

There are several \textit{metrics} to quantify token importance including recency, aggregate attention scores, and others listed in Table~\ref{tab:related_work_table}.
Token Sparsity methods use these metrics to guide token eviction or retrieval decisions.
There is an important interplay between methods and metrics, with trade-offs associated with eviction or grouped coarse grained token-access patterns.
To address this, we propose a novel \textit{learned} metric of token importance, which provides fine-grained estimates of token importance, and we provide a system called \textbf{TokenButler} that makes use of it. Our contributions are summarized as:

\begin{itemize}
    \item A lightweight predictor ($<1\%$ parameter overhead) that learns token importance via attention distillation.
    \item A synthetic co-referential benchmark exposing failures of existing methods at contexts as small as 300 tokens; TokenButler achieves near-oracle accuracy.
    \item Competitive or superior results on RULER and LongBench at 64K context after using max length of 1K for distillation, and up to 7.6$\times$ latency reduction over Dense Attention with CPU offloading.
    \item A \emph{prediction interval} scheme that amortizes predictor cost by $16\times$, with \emph{neighbor fetching} to maintain accuracy within $\approx$1.1\%, bringing on-GPU speedup over Dense Attention to 1.6$\times$.
\end{itemize}

\section{Related Work}

Transformers exhibit strong contextual behavior: head and neuron importance depends heavily on the input query.~\citet{pmlr-v202-liu23am_dejavu, akhauri-etal-2024-shadowllm} train small predictors to prune neurons and heads on a per-query basis using magnitude or gradient-based \textit{metrics}. We draw the same distinction between importance \textit{metrics} and the pruning \textit{methods} that act on them.

\begin{table}
  \centering
  \vspace{-1cm}
  \small
  \begin{tabular}{lp{0.65\linewidth}}
    \toprule
    \textbf{Method}    & \textbf{Metric} \\
    \midrule
    StreamingLLM           & Recency-based sliding window \\
    H\textsubscript{2}O & Attention Score for Token Eviction \\
    SnapKV              & Pooled Attention Score over a Fixed Window for Token Eviction \\
    Quest               & Query product with Per-Page Min–Max Token Magnitudes for Page Loading \\
    TokenButler         & Predicted Importance for Fine-Grained Token Loading \\
    \bottomrule
  \end{tabular}
  \caption{Metrics for token importance}
  \label{tab:related_work_table}
  \vspace{-14pt}
\end{table}

This contextual behavior applies to token importance by design, as the attention mechanism explicitly captures tokens relevant to a query. However, while \textit{methods} to prune heads are simpler, as there is a fixed number of heads, \textit{methods} to prune tokens are more expensive to realize. Specifically, for a transformer with $N$ layers and $H$ heads per-layer and $L$ past-tokens, every head has to decide which \textit{subset} $S$ of $L$ tokens are the most important at every decode step. This implies that any given \textit{metric} has to be calculated for $N\times H\times L$ tokens, at \textit{every decode step}.

As presented in Table \ref{tab:related_work_table}, there have been significant efforts towards \textit{co-designing} metrics with methods of token sparsity.
The simplest methods are \textit{purely static strategies}, StreamingLLM~\citep{xiaoefficientstreamllm} relies on \textit{recency} as a metric of token importance, with a sliding-local-window plus initial anchor tokens attention to fix a KV-Cache budget.
More recently, methods like $H_2O$~\citep{zhang2023h2o} and SnapKV~\citep{li2024snapkv} avoid na\"ive sparsification of tokens, and instead rely on attention scores to permanently evict low-importance tokens.
This can be a major limitation when tasks require synthesizing or reasoning over information distributed across the context~\citep{vodrahalli2024michelangelolongcontextevaluations}, as a token that becomes important later in the decoding stage may be evicted due to its low importance at the current step and low KV-Budget. Furthermore, these methods typically rely on accumulating attention scores over a sliding window to determine long-term importance. This accumulation is inherently biased towards tokens that are frequently attended to in the short term (high frequency), potentially penalizing 'rare-event' tokens that are crucial for answering specific future queries but possess low aggregate attention scores (high utility, low frequency).
To alleviate this issue, \textit{Adaptive Dynamic Strategies} such as Quest~\citep{tang2024quest} preserve all tokens, and dynamically decide which subset of tokens to fetch for a given query. Instead of calculating full attention scores to ensure the most important tokens are fetched (which can be prohibitively expensive), Quest relies on paging, preserving all tokens in paged memory, and selectively fetches important pages.
To determine page importance, the dot product of query with min-max token values within a page is used as a proxy.
This reduces memory bandwidth but does not optimize memory footprint. Furthermore, its sparsity is limited to the granularity of pages limiting its effectiveness in more challenging co-referential tasks as we will show.
TokenSelect~\citep{wu2024tokenselect} also preserves all tokens and selects the important ones based on the dot product between queries and keys but it intelligently avoids doing that with every query based on the cosine similarity between different queries. However, this method incurs a high overhead due to the need of performing dot products with a high dimension.

These attention-based metrics are tightly coupled to their methods, requiring eviction, paging, or expensive scoring. By contrast, \textit{TokenButler} learns a lightweight predictor (${\sim}$1\% of the LLM) that cheaply approximates token-level attention logits via low-dimensional QK projections, preserving fine-grained per-token control.

Concurrent to our work, DeepSeek-V3.2~\citep{deepseekv32} introduces a \emph{Lightning Indexer}, a lightweight FP8 scoring module that selects important tokens for sparse attention. While the high-level motivation is similar, the two approaches differ in important ways. First, DeepSeek Sparse Attention requires extensive continued training of the full model (${\sim}$944B tokens) to adapt to the sparse regime, whereas TokenButler trains only a small external predictor on top of a frozen pretrained LLM, making it applicable to any off-the-shelf model at minimal cost. Second, the Lightning Indexer must recompute index scores at every decode step and every layer, whereas TokenButler supports \textit{prediction intervals} with neighbor fetching (Section~\ref{sec:prediction_interval}), amortizing predictor cost across multiple decode steps with negligible accuracy loss.

\section{Methodology}

\begin{figure*}[t]
    \centering
    \includegraphics[width=1\linewidth]{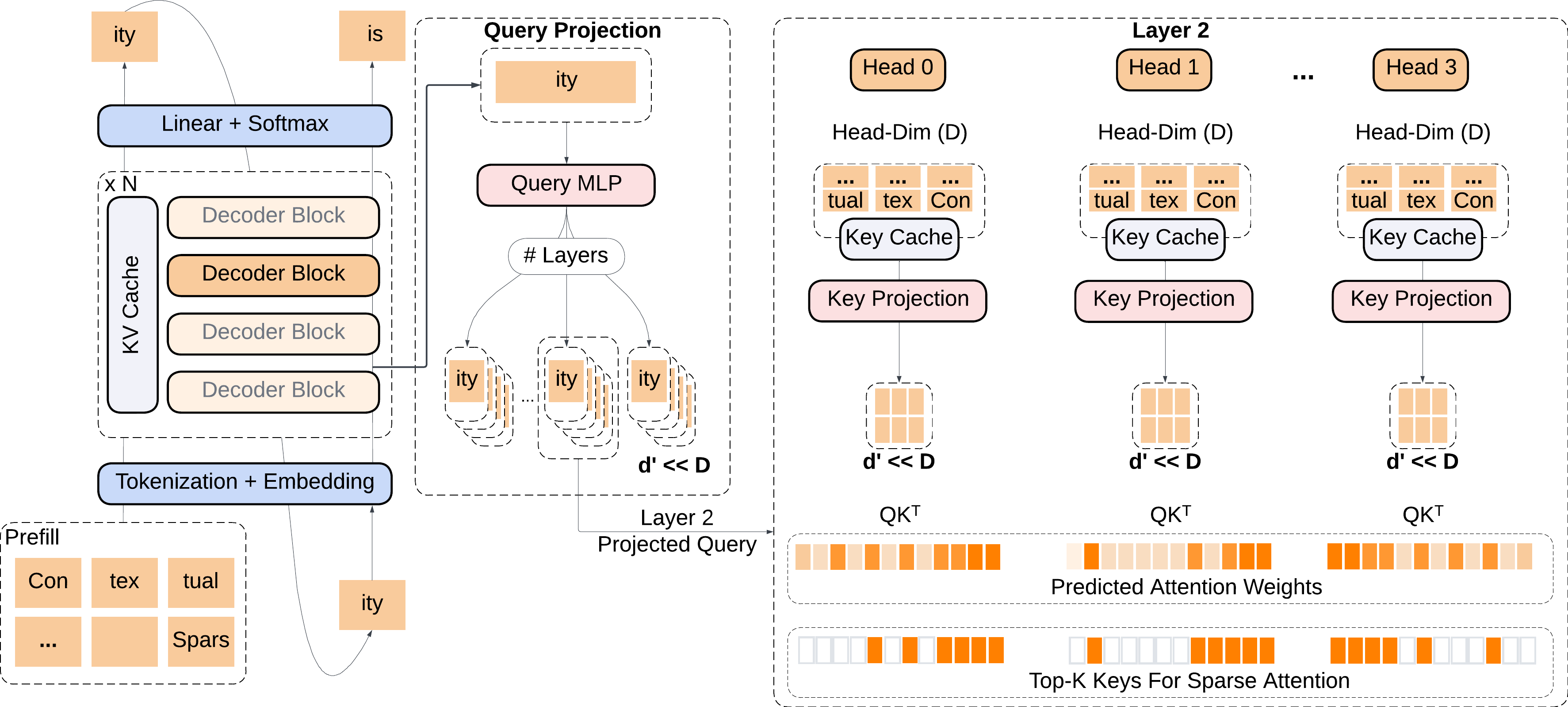}
    \caption{TokenButler predicts low-dimensional importance queries at fixed depth strides (producer layers) and combines them with a learned projection of the real KV-cache keys at each consumer layer to score and select tokens under a fixed budget. Training distills the masked causal attention distribution of the frozen LLM, and inference applies per-token selection while preserving sink tokens and local window tokens.}
    \label{fig:pred_train_fig}
    \vspace{-2pt}
\end{figure*}

TokenButler (Figure~\ref{fig:pred_train_fig}) uses a single LLM layer's output to predict token importance across multiple subsequent layers ahead of time, trained by distilling the LLM's own attention distributions.

\subsection{TokenButler Design}
\label{sec:predictor_design}

TokenButler operates at a fixed depth stride $G$ (\texttt{producer\_frequency}), designating \emph{producer} layers at indices $0, G, 2G, \dots$.
Each producer takes hidden states $\mathbf{H}\in\mathbb{R}^{B\times L\times E}$ and predicts low-dimensional importance queries for the next $G$ \emph{consumer} layers; consumer layer $\ell$ uses slot $(\ell{-}1)\bmod G$.

\paragraph{Query prediction.}
A lightweight MLP produces $G$ slot-specific query vectors per head:
\[
\mathbf{Q}_{\mathrm{imp}} = f_{\theta}(\mathrm{LN}(\mathbf{H})) \in \mathbb{R}^{(BH)\times G \times L \times d'},
\]
where $d' \ll E$ is the interaction dimension.

\paragraph{Key-cache projection.}
For each layer $\ell$, we project the \emph{real KV-cache keys} into the same $d'$-dimensional space via a learned matrix $\mathbf{W}^{(\ell)}_{\mathrm{K}}\in\mathbb{R}^{D\times d'}$:
\[
\mathbf{K}^{(\ell)}_{\mathrm{imp}} = \mathbf{K}^{(\ell)} \mathbf{W}^{(\ell)}_{\mathrm{K}} \in \mathbb{R}^{B\times H_{\mathrm{kv}}\times L \times d'}.
\]
This anchors scoring to the model's true cached keys while keeping computation low-dimensional.

\subsection{Training Objective}
\label{sec:training}

We freeze the base LLM and train only TokenButler parameters (the query MLPs and key-cache projection matrices).
Given an input sequence, we compute teacher attention logits from the frozen model for each layer and head, and compute TokenButler's student logits using predicted importance queries and projected KV-cache keys.
Crucially, both teacher and student logits include the same causal/padding mask prior to normalization.

We distill the \emph{masked causal attention distribution}.
Let $\mathbf{A}_{\mathrm{true}}$ be the teacher logits (after adding the attention mask) and $\mathbf{A}_{\mathrm{pred}}$ be the corresponding TokenButler logits.
We define:
\[
\mathbf{P} = \mathrm{softmax}(\mathbf{A}_{\mathrm{true}}), \qquad
\mathbf{Q} = \mathrm{softmax}(\mathbf{A}_{\mathrm{pred}}),
\]
and minimize a cross-entropy distillation loss:
\[
\mathcal{L}_{\mathrm{CE}} = -\mathbb{E}\left[\sum_{k} \mathbf{P}_{k}\log(\mathbf{Q}_{k})\right].
\]

To make training efficient at long context lengths, we subsample a fixed number of query positions per sequence: we draw most loss rows from the late-context region and always include the final token, reducing the auxiliary loss cost from $\mathcal{O}(L^2)$ to $\mathcal{O}(RL)$ with $R \ll L$.

\subsection{Inference Setup}
\label{sec:inference}

Building on the critical role of initial and recent tokens~\cite{xiaoefficientstreamllm}, we partition the KV-cache into three contiguous buffers: a \textbf{Sink Buffer} (first $S$ tokens, always retained), a \textbf{Local Window Buffer} (circular buffer of the most recent $N$ tokens), and an \textbf{Important Buffer} (dynamically populated by TokenButler). This layout keeps attention kernel accesses contiguous, avoiding fragmented memory gathers.

To minimize the overhead of projecting high-dimensional keys ($D$) into the low-dimensional importance space ($d'$), we leverage the temporal locality of the Local Window Buffer. Because the standard attention kernel always includes the local window densely, any newly generated token is automatically attended to without requiring an importance score for the first $N$ steps of its "life". This allows us to defer the $d'$-dimensional projection until the token is ready to be evicted from the window and evaluated for its long-term relevance. Consequently, projections are performed in batches of $N$ tokens rather than at every decode step. Starting from a post-prefill state where all existing keys are projected, we allow the local window to fill over $N$ steps. Only when the buffer head returns to the start of the circular queue do we project the $N$ most recent keys in a single batch and append them to the predictor's search space. This batching significantly improves system throughput by utilizing high-bandwidth memory (HBM) more efficiently.

We similarly batch the remaining predictor operations. Once the input of the predictor at a producer layer is ready, the system performs query prediction, importance scoring, and the migration of selected KV-pairs into the Important Buffer for all the subsequent consumer layers. Doing these operations in batches improves the efficiency of the system.

\subsection{Prediction Interval and Neighbor Fetching}
\label{sec:prediction_interval}

While TokenButler's predictor is lightweight, its per-step cost (importance query prediction, score computation against all projected keys, and KV gather) accumulates over long generations. We introduce \emph{prediction interval with neighbor fetching}, an optimization that reduces predictor overhead by up to $N{\times}$ with minimal accuracy loss.

\paragraph{Prediction interval ($i{=}N$).} Instead of running the predictor at every decode step, we invoke it once every $N$ steps. On prediction steps, the full pipeline executes: predictor forward pass, score computation, top-$B$ selection, and KV gather into the Important Buffer. On the intervening $N{-}1$ steps, attention reuses the \emph{stale} sparse selection from the last prediction. The Local Window Buffer is updated every step regardless, ensuring that the most recent $W$ tokens are always available. Over a generation of $T$ tokens, this reduces predictor invocations from $T$ to approximately $T/N$.

\paragraph{Neighbor fetching.} Reusing stale selections risks missing tokens whose importance has shifted since the last prediction. To mitigate this, we expand each selected token to include its spatial neighbor, leveraging the observation that important information often spans consecutive tokens (e.g., multi-token entities or reasoning chains). We use a cluster-aware algorithm: consecutive selected indices form clusters, and each token's neighbor is placed just past its cluster boundary to maximize coverage. This yields $2B$ unique positions per prediction step, and the sparse buffer is sized accordingly. By doubling the spatial coverage, neighbor fetching provides protection against importance drift between prediction refreshes: if the importance landscape shifts slightly, spatial neighbors of the originally selected tokens are likely to cover the newly important positions. All results with $i > 1$ employ neighbor fetching.

\section{Experiments}
\label{sec:sec4}
Across all experiments, we use the same predictor architecture: a producer every \(G{=}4\) layers, interaction dimension \(d'{=}16\), and a two-layer MLP with hidden size 512. The base LLM is frozen; only the predictor is trained (\(\texttt{lr}{=}10^{-3}\)) on a mixture of web, code, and long-context QA data at sequence length 1K. Since the predictor projects the key-cache directly, it generalizes to 64K contexts without long-context fine-tuning. Full training details are in Appendix~\ref{app:training_details}.

\paragraph{Token Selection Policy}
TokenButler produces a per-layer token-importance score over the KV-cache tokens and converts it into a binary keep/drop decision under a fixed budget.
We define a set of \emph{candidate} tokens eligible for pruning by excluding: (i) a prefix of sink tokens of length $S$ (always retained), and (ii) a  local window tail of length $W$ (always retained).
Among the remaining candidates, we apply a fixed sparsity rule (\\texttt{xtok}), retaining the top-$x$ tokens by predicted importance.

As is standard, we apply sparsification only after the model has observed a dense prefix of the sequence: the prefill pass remains dense and the first decode token is also computed densely; TokenButler pruning is applied from the subsequent decode steps onward.

\subsection{Evaluation On a Synthetic Task for Token Retrieval}

\begin{figure*}[ht]
    \centering
    \includegraphics[width=\linewidth]{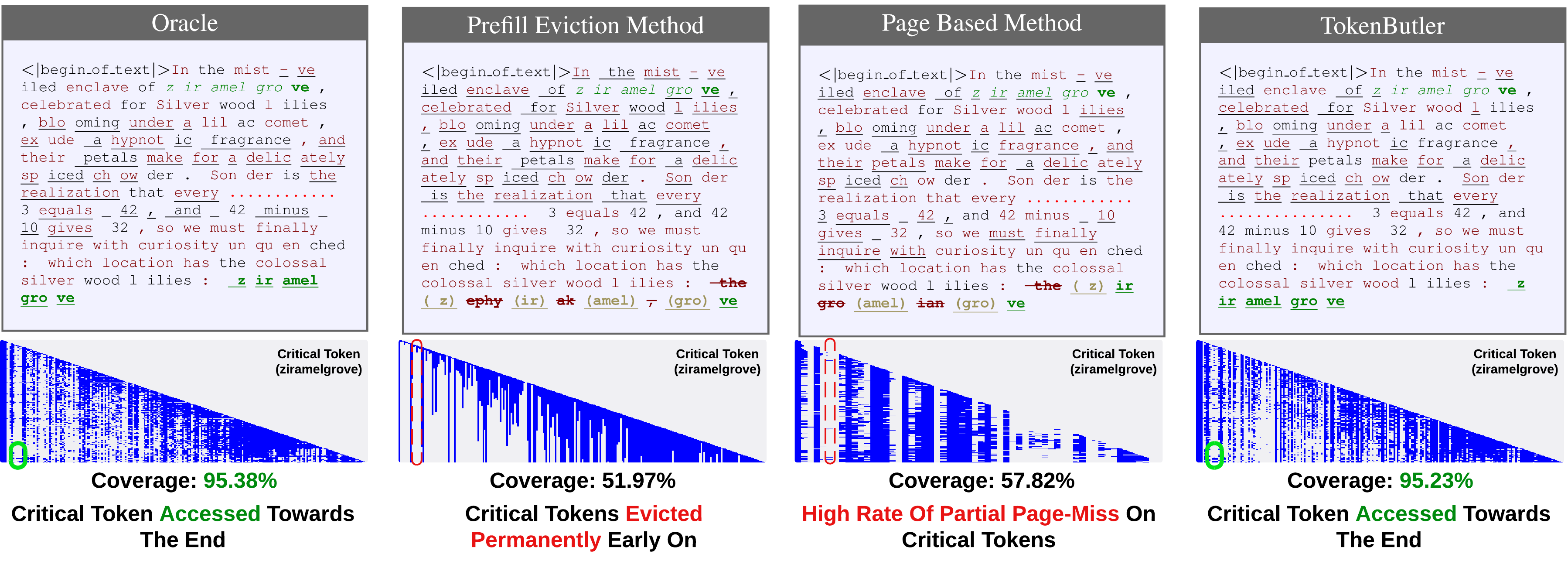}
    \caption{Sample behavior of different KV-Sparsity methods on our synthetic co-reference resolution task. \comment{The tokens that are accessed at the decode step for location \textit{ziramelgrove} are underlined. The tokens that were mis-predicted at the last step of the decode simulation are in red. The positions of the location is in green, and if the final location is mis-predicted, that token is striked-out in red.} TokenButler outperforms prefill eviction and page-based methods that have clear failure modes due to permanently dropping tokens, or fetching tokens with page-size granularity respectively.}
    \label{fig:missrate_withmap}
\end{figure*}

\begin{figure*}[t]
\centering

\begin{minipage}[t]{0.40\textwidth}
\vspace{0pt} %

\begin{tcolorbox}[
    colframe=black!60,        %
    colback=blue!5,          %
    coltitle=white,          %
    sharp corners,           %
    boxrule=0.8pt,           %
    left=4pt, right=4pt, top=6pt, bottom=6pt,
    fontupper=\ttfamily,     %
    before skip=8pt,         %
    after skip=8pt,          %
    width=\columnwidth,       %
    title={Synthetic Benchmark Template and Sample} %
]
\fontsize{7}{8}\selectfont
<\textcolor{purple}{contextual lead}>
\textbf{\textcolor{teal}{<location>}}
<\textcolor{black}{philosophical statement}>
<\textcolor{red}{culinary statement}>
<\textcolor{brown}{math problem}>
<\textbf{\textcolor{purple}{location prelude}}>
\hspace{1em}\textbf{\textcolor{teal}{<<location>>}}

\noindent\rule{\linewidth}{0.4pt}

\fontsize{7}{8}\selectfont
\textcolor{purple}{Shrouded in fog, .... place is:}
\textbf{\textcolor{teal}{wraithspire.}}
\textcolor{black}{In the ... wisdom to sprout.}
\textcolor{red}{Savor the ...  home-cooked love.}
\textcolor{brown}{If we compute ... gives us 16.}
\textbf{\textcolor{purple}{Which location ... up the shore?}}
\hspace{1em}\textbf{\textcolor{teal}{<<wraithspire>>}}
\end{tcolorbox}
\end{minipage}
\hfill
\begin{minipage}[t]{0.58\textwidth}
\vspace{8pt}

\centering
\resizebox{\linewidth}{!}{%
\begin{tabular}{l|cc|cc|cc|cc}
  \toprule
  \multirow{2}{*}{\textbf{Llama}} & \multicolumn{2}{c|}{\textbf{Oracle}} & \multicolumn{2}{c|}{\textbf{Token Eviction}} & \multicolumn{2}{c|}{\textbf{Page-Based}} & \multicolumn{2}{c}{\textbf{TokenButler}} \\
  & Acc. & Cov. & Acc. & Cov. & Acc. & Cov. & Acc. & Cov. \\
  \midrule
  1B   & 49.00 & 84.32 & 1.00  & 32.50 & 0.00    & 19.78 & \textbf{48.94} & \textbf{84.02} \\
  3B   & 81.00 & 95.38 & 10.00 & 51.97 & 6.00 & 57.82 & \textbf{80.20} & \textbf{95.23} \\
  8B   & 77.00 & 93.47 & 3.00  & 37.50 & 0.00    & 46.98 & \textbf{76.17} & \textbf{92.59} \\
  \bottomrule
\end{tabular}
}
\captionof{table}{Accuracy and coverage (\%) of different KV-sparsity methods on our synthetic dataset. TokenButler outperforms eviction and page-based methods, and approaches Oracle performance.}
\label{tab:metrics_comparison}

\end{minipage}

\end{figure*}

We evaluate TokenButler on a difficult synthetic task inspired by Multi-Round Co-reference Resolution~\citep{vodrahalli2024michelangelolongcontextevaluations}, using concise sequences (\(<512\) tokens). The model must recall a fictional location mentioned in a \textit{contextual lead}, then referenced again after several distracting statements. By the time the location needs to be mentioned again after the location prelude, several tokens may have intervened, making it likely that the location tokens may have been evicted. Coarse-grained retrieval schemes risk not finding the entire location as it may be split across pages. This setup mimics conversation-like scenarios. It is especially challenging for token sparsity methods, since prematurely discarding or overlooking the location tokens can irreversibly break the final reference, leading to incorrect or incomplete retrieval of the location name. Our data generation procedure is further detailed in Appendix ~\ref{appx:synthetictask}

Since every head may evict tokens based on their importance, we present the attention map for the first head of the 3rd layer (a random choice) in Figure~\ref{fig:missrate_withmap}.
We observe there as well as in Table \ref{tab:metrics_comparison} that (i)~prefill eviction methods, e.g. \(\mathrm{H_2O}\), have low accuracy because they permanently evict older tokens (the location name) once new context is being decoded. (ii)~page-based methods, e.g. Quest, \textbf{very often} lose part of the location name if it straddles a page boundary in this context-dense example.
\textit{Coverage} counts the fraction of correctly-predicted location tokens (e.g., 3/4 tokens correct yields 0.75 coverage but 0 accuracy). Eviction and page-based methods recover 30-50\% of tokens but rarely all of them, leading to low accuracy (Table~\ref{tab:metrics_comparison}).

\begin{table*}[ht]
    \centering
    \vspace{5pt}
    \resizebox{\textwidth}{!}{
    \begin{tabular}{cr|rrrrrrrrrrr|r}
    \toprule
    Method & K.V. BW ($\times \downarrow$) & NarQA & Qasper & MFQA & HotpotQA & 2WikiMQA & Musique & GovRpt & QMSum & MultiNews & SAMSum & PassRet & Avg. \\
    \midrule
    \multicolumn{14}{c}{\textbf{Llama-3.1-8B-Instruct}} \\ 
    \midrule
    Dense & 1.00 & 30.64 & 46.64 & 55.42 & 58.17 & 47.76 & 28.23 & 34.51 & 25.61 & 25.43 & 23.51 & 100.00 & 43.27 \\
    \midrule
    MiniCache & 1.30 & 18.11 & 20.41 & 27.51 & 38.80 & 21.37 & 20.19 & 21.09 & 20.85 & 19.60 & 22.75 & 98.00 & 29.88 \\
    StreamingLLM & 8.00 & 26.25 & 29.32 & 34.56 & 49.99 & 42.47 & 20.92 & 26.29 & 21.90 & 22.63 & 23.87 & 93.50 & 35.61 \\
    SnapKV & 8.00 & 30.34 & 45.56 & 53.62 & 57.92 & 47.50 & 26.96 & 28.99 & 25.33 & 23.95 & 24.91 & 99.50 & 42.23 \\
    PyramidKV & 8.00 & 31.83 & 44.36 & 52.56 & 56.38 & 48.35 & 27.23 & 28.63 & 25.06 & 23.49 & 24.80 & 99.50 & 42.02 \\
    KIVI & 7.10 & 29.84 & 41.49 & 51.28 & 57.27 & 42.54 & 27.54 & 33.28 & 25.01 & 23.90 & 25.88 & 98.17 & 41.47 \\
    Single SVD & 8.40 & 15.52 & 38.36 & 37.14 & 30.04 & 19.70 & 22.06 & 19.52 & 22.35 & 18.91 & 22.82 & 85.00 & 30.13 \\
    xKV & 8.03 & 32.85 & 45.62 & 54.69 & 51.74 & 36.38 & 28.33 & 31.32 & 24.49 & 22.61 & 26.26 & 100.00 & 41.30 \\
    \rowcolor{lightmintbg} \textbf{TokenButler} & \textbf{8.00} & \textbf{30.73} & \textbf{48.86} & \textbf{56.45} & \textbf{57.77} & \textbf{50.82} & \textbf{28.34} & \textbf{34.34} & \textbf{25.30} & \textbf{26.91} & \textbf{29.22} & \textbf{100.00} & \textbf{44.43} \\

    \midrule
    \multicolumn{14}{c}{\textbf{Qwen2.5-7B-Instruct-1M}} \\ 
    \midrule
    Dense & 1.00 & 29.21 & 43.78 & 48.58 & 60.92 & 53.29 & 33.68 & 33.23 & 23.24 & 23.53 & 43.21 & 100.00 & 44.79 \\
    \midrule
    MiniCache & 1.30 & 7.98 & 16.47 & 9.42 & 21.78 & 15.46 & 10.55 & 8.83 & 7.94 & 3.84 & 11.32 & 5.00 & 10.78 \\
    StreamingLLM & 8.00 & 24.13 & 34.99 & 27.47 & 47.55 & 45.13 & 21.44 & 26.76 & 19.34 & 21.15 & 44.11 & 29.00 & 31.01 \\
    SnapKV & 8.00 & 28.68 & 44.51 & 47.43 & 60.88 & 51.91 & 32.17 & 29.95 & 23.05 & 21.79 & 45.67 & 100.00 & 44.19 \\
    PyramidKV & 8.00 & 29.93 & 41.29 & 47.21 & 60.16 & 51.98 & 32.79 & 27.21 & 22.88 & 19.86 & 45.00 & 100.00 & 43.48 \\
    KIVI & 7.10 & 26.43 & 35.78 & 38.40 & 46.17 & 45.08 & 23.79 & 24.62 & 22.41 & 16.98 & 39.52 & 63.50 & 34.79 \\
    Single SVD & 8.40 & 27.88 & 41.84 & 47.66 & 52.73 & 48.76 & 28.46 & 28.71 & 22.24 & 21.18 & 33.18 & 65.00 & 37.97 \\
    xKV & 8.03 & 28.78 & 42.98 & 47.43 & 58.79 & 52.42 & 32.69 & 33.02 & 23.67 & 23.18 & 38.62 & 98.00 & 43.60 \\
    \rowcolor{lightmintbg}\textbf{TokenButler} & \textbf{8.00} & \textbf{29.39} & \textbf{47.27} & \textbf{52.50} & \textbf{59.67} & \textbf{56.01} & \textbf{34.58} & \textbf{33.32} & \textbf{23.14} & \textbf{24.04} & \textbf{35.09} & \textbf{100} & \textbf{45.00} \\

    \bottomrule
    \end{tabular}
    }
    \caption{Comparing TokenButler with KV-Cache sparsity, quantization and low-rank compression methods from prior research.}
    \label{tab:longbench}
\end{table*}

\subsection{Long Context Evaluation}

\begin{table*}[ht]
    \centering
    \resizebox{\textwidth}{!}{
    \begin{tabular}{cr|rrrrrrrrrr|r}
    \toprule
    Method & K.V. BW ($\times \downarrow$) & N-S1 & N-S2 & N-MK1 & N-MK2 & N-MQ & N-MV & QA-1 & QA-2 & VT & FWE & Avg. \\
    \midrule
    \multicolumn{13}{c}{\textbf{Llama-3.1-8B-Instruct}} \\
    \midrule
    Dense & 1.00 & 100.00 & 100.00 & 98.96 & 97.92 & 98.96 & 97.66 & 83.33 & 59.38 & 97.29 & 85.42 & 91.89 \\
    \midrule
    KVzip & 8.00 & 100.00 & 100.00 & 8.33 & 70.83 & 52.60 & 78.12 & 62.50 & 58.33 & 73.96 & 75.35 & 68.00 \\
    SnapKV & 8.00 & 100.00 & 100.00 & 98.96 & 94.79 & 100.00 & 97.66 & 83.33 & 58.33 & 95.00 & 68.75 & 89.68 \\
    Quest & 8.00 & 90.75 & 90.63 & 96.88 & 87.50 & 94.27 & 85.42 & 83.33 & 57.29 & 77.71 & 81.94 & 84.57 \\
    xKV & 8.03 & 100.00 & 96.88 & 97.92 & 97.92 & 96.09 & 96.62 & 78.13 & 56.25 & 86.67 & 78.47 & 88.50 \\
    \rowcolor{lightmintbg}\textbf{TokenButler} & \textbf{8.00} & \textbf{100.00} & \textbf{100.00} & \textbf{100.00} & \textbf{96.88} & \textbf{98.96} & \textbf{94.27} & \textbf{83.33} & \textbf{57.29} & \textbf{91.46} & \textbf{77.08} & \textbf{89.93} \\
    \bottomrule
    \end{tabular}
    }
    \caption{RULER evaluation on Llama-3.1-8B-Instruct at 64K context length. K.V. BW ($\times \downarrow$) refers to the reduction in key-value cache access bandwidth.}
    \label{tab:ruler_st}
\end{table*}

\begin{table*}[ht]
    \centering
    \resizebox{\textwidth}{!}{
    \begin{tabular}{cr|rrrrrrrrrr|r}
    \toprule
    Method & K.V. BW ($\times \downarrow$) & N-S1 & N-S2 & N-MK1 & N-MK2 & N-MQ & N-MV & QA-1 & QA-2 & VT & FWE & Avg. \\
    \midrule
    \multicolumn{13}{c}{\textbf{Qwen2.5-7B-Instruct-1M}} \\ 
    \midrule
    Dense & 1.00 & 100.00 & 100.00 & 100.00 & 100.00 & 100.00 & 95.83 & 84.38 & 60.42 & 90.63 & 86.81 & 91.81 \\
    \midrule
    MiniCache & 1.30 & 26.04 & 0.00 & 0.00 & 0.00 & 0.00 & 0.00 & 12.50 & 14.58 & 0.42 & 3.47 & 5.70 \\
    Single SVD & 8.40 & 100.00 & 97.92 & 96.88 & 98.96 & 97.40 & 91.15 & 64.58 & 56.25 & 73.75 & 61.46 & 83.84 \\
    xKV & 8.03 & 100.00 & 100.00 & 100.00 & 98.96 & 100.00 & 90.63 & 80.21 & 58.33 & 82.08 & 81.94 & 89.22 \\
    StreamingLLM & 8.00 & 15.63 & 12.50 & 12.50 & 9.38 & 14.84 & 17.71 & 46.88 & 43.75 & 13.13 & 89.24 & 27.56 \\
    SnapKV & 8.00 & 100.00 & 96.88 & 97.92 & 31.25 & 95.31 & 83.07 & 84.38 & 59.38 & 91.25 & 80.56 & 82.00 \\
    PyramidKV & 8.00 & 100.00 & 93.75 & 96.88 & 16.67 & 90.37 & 80.73 & 84.38 & 59.38 & 89.17 & 76.39 & 78.77 \\
    KIVI & 7.10 & 0.00 & 2.08 & 3.13 & 13.54 & 0.00 & 0.78 & 48.96 & 43.75 & 36.46 & 40.63 & 18.93 \\
    \rowcolor{lightmintbg}\textbf{TokenButler} & \textbf{8.00} & \textbf{100.00} & \textbf{100.00} & \textbf{100.00} & \textbf{96.88} & \textbf{99.47} & \textbf{90.63} & \textbf{84.38} & \textbf{58.33} & \textbf{89.17} & \textbf{82.64} & \textbf{90.15} \\
    \bottomrule
    \end{tabular}
    }
    \caption{TokenButler vs. methods which sparsify or compress the KV-Cache.}
    \label{tab:ruler_st_only}
\end{table*}

We evaluate TokenButler under a fixed decode-time KV-access budget corresponding to an effective \(8\times\) compression at 64K context. Specifically, at each decode step we always retain a small set of \emph{sink} prefix tokens (128) and a \emph{sliding-window} of most-recent tokens (local window = 256), and allocate the remaining budget to a query-dependent, per-token selection from the rest of the context so that the total number of tokens accessed per layer is capped at \(\approx 8\)K.

Table \ref{tab:ruler_st} shows the results of RULER benchmark. TokenButler achieves competitive performance on Llama-3.1-8B-Instruct, outperforming baselines like Quest and KVzip while matching stronger methods like SnapKV and xKV. We also evaluate the \texttt{Qwen-2.5-7B-Instruct-1M} model in Table \ref{tab:ruler_st_only}, where TokenButler achieves the best average score among all methods. Both evaluations are done at a context-length of 64K.

\paragraph{LongBench} ~\cite{bai2024longbenchbilingualmultitaskbenchmark}.
We evaluate the \texttt{Llama-3.1-8B-Instruct} and \texttt{Qwen2.5-7B-Instruct-1M} models across a wide range of baselines that alleviate KV-Cache bandwidth requirements. Specifically, we evaluate KV-compression methods such as PyramidKV~\cite{cai2024pyramidkv}, SingleSVD, MiniCache~\cite{liu2024minicache} and xKV~\cite{chang2025xkv}, quantization with KIVI~\cite{liu2024kivi} and sparsity with StreamingLLM~\cite{xiaoefficientstreamllm}, SnapKV~\cite{li2024snapkv} and TokenButler. We find that TokenButler offers better accuracy than prior methods from Table \ref{tab:longbench}.

\begin{figure}[t]
    \centering
    \begin{subfigure}[t]{0.48\columnwidth}
        \centering
        \includegraphics[width=\linewidth]{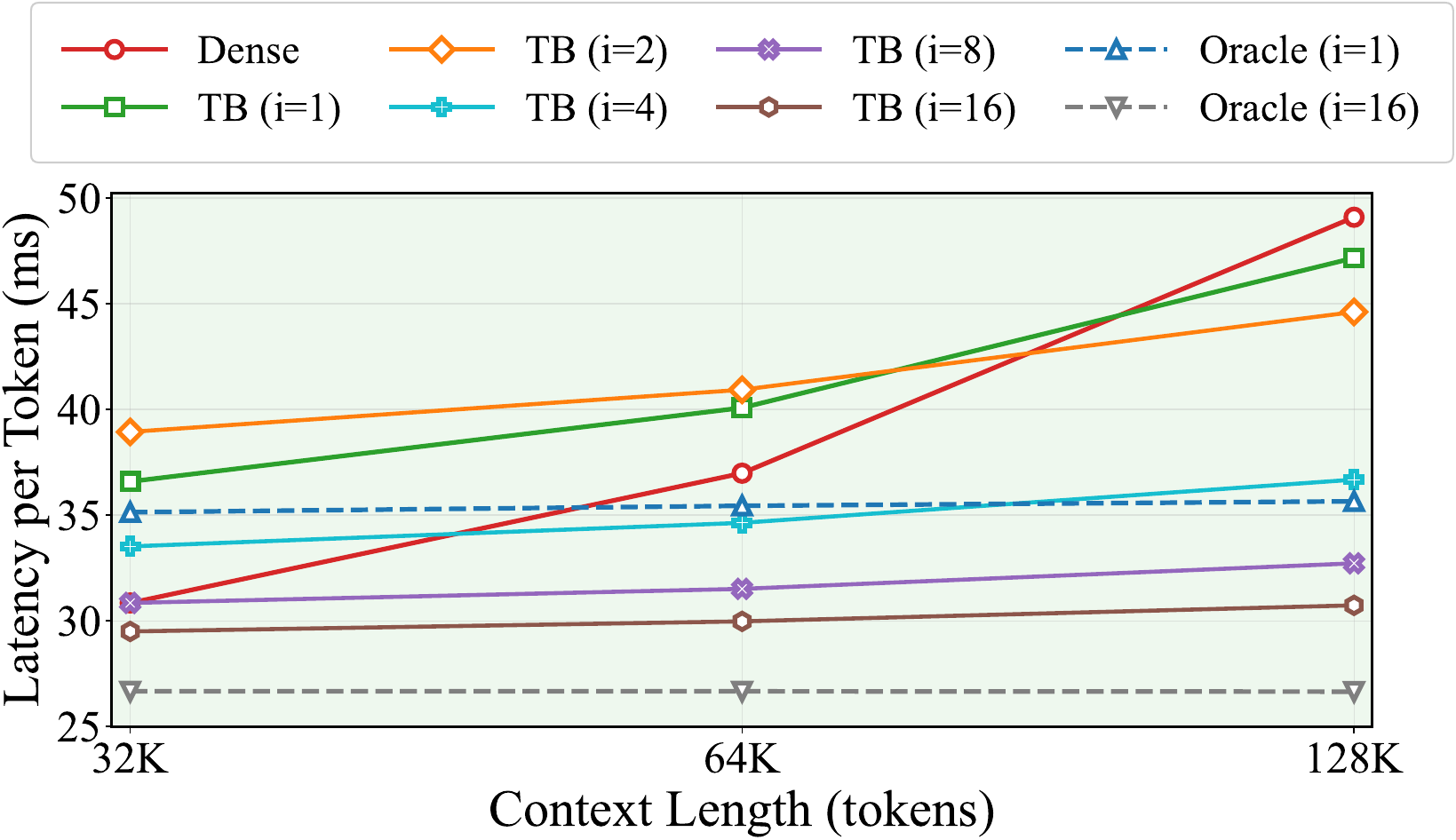}
        \caption{\footnotesize{Using GPU Memory. TB = TokenButler. i is prediction interval}}
        \label{fig:interval_latency}
    \end{subfigure}
    \hfill
    \begin{subfigure}[t]{0.48\columnwidth}
        \centering
        \includegraphics[width=\linewidth]{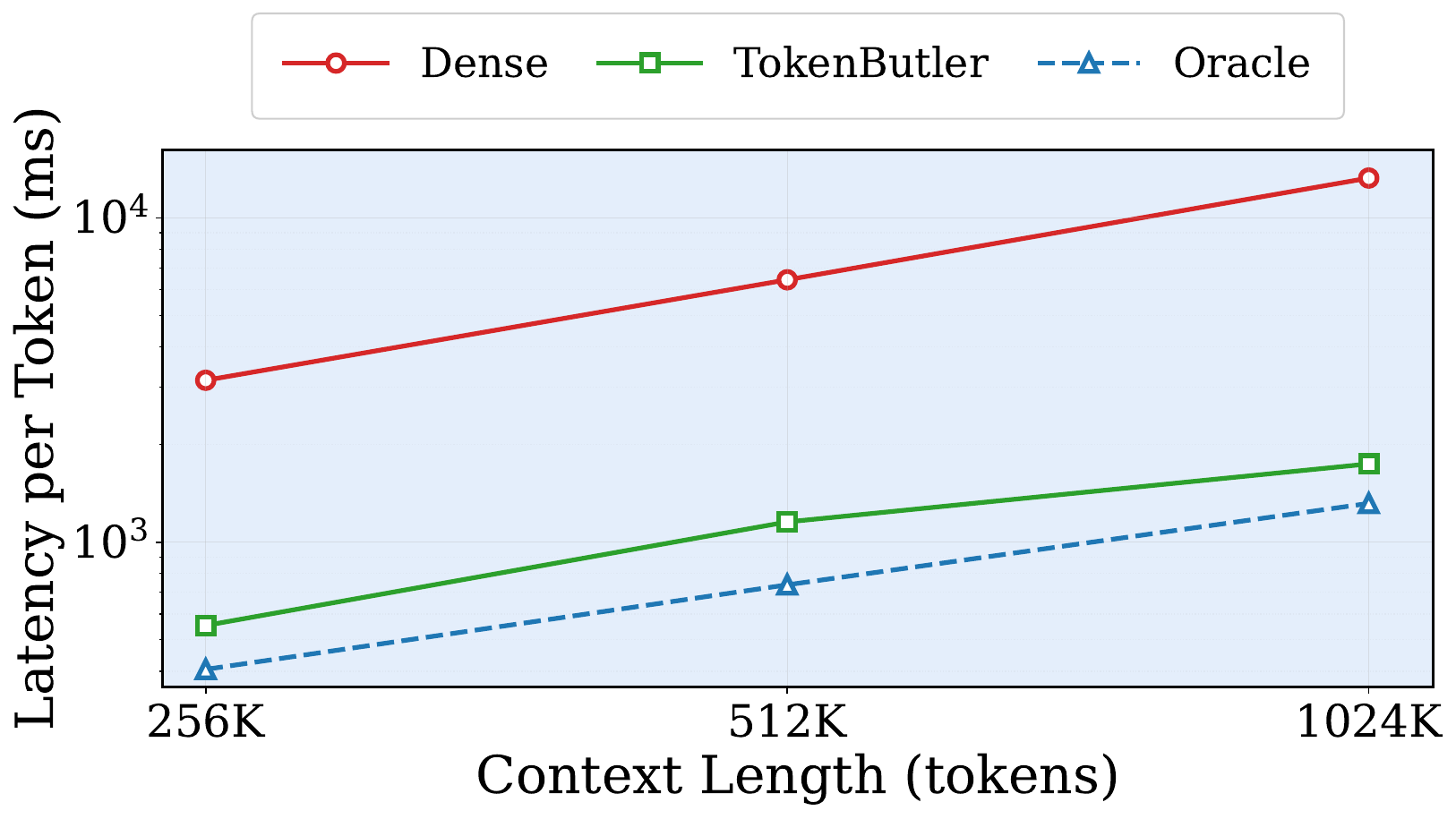}
        \caption{Offloaded to CPU.}
        \label{fig:decoding_efficiency_8b}
    \end{subfigure}
    \caption{Decoding efficiency on \texttt{Llama-3.1-8B-Instruct}. Sparse Token Budget is set to 8K.}
    \label{fig:efficiency_combined}
\end{figure}

\paragraph{Reasoning models}
\begin{wrapfigure}{r}{0.45\columnwidth}
    \centering
    \includegraphics[width=\linewidth]{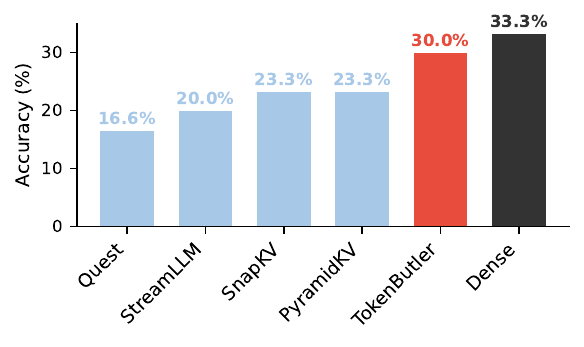}
    \caption{AIME24 accuracy.}
    \label{fig:aime24_reasoning}
    \vspace{-20pt}
\end{wrapfigure}
The emphasis on long decode has increased, as reasoning models must emit long chain-of-thought (CoT) traces before arriving at an answer. These generated CoTs can significantly slow down decoding by stressing memory bandwidth with excessive KV-token loading. We train TokenButler on \texttt{DeepSeek-R1-Distill-Llama-8B}~\cite{deepseekai2025deepseekr1incentivizingreasoningcapability} and evaluate on AIME24 with a decode budget of 8K tokens. Across all methods, we fix a maximum KV access budget of 2179 tokens per step (32 sink tokens and 128 local window tokens). Since most AIME24 problems require fewer than 160 prefill tokens, this setting is decode-heavy with a relatively small prefill.
Figure~\ref{fig:aime24_reasoning} summarizes accuracy: TokenButler attains 30.0\% accuracy, approaching dense decoding (33.3\%) and outperforming other KV-sparsity baselines under the same budget.

\subsection{Efficiency}
\label{sec:interval_results}

We measure per-token decode latency on \texttt{Llama-3.1-8B-Instruct} (Nvidia A6000, budget = 8K tokens) using prediction interval with neighbor fetching (Section~\ref{sec:prediction_interval}). For contexts exceeding 128K we employ CPU offloading; we also report an Oracle (zero-overhead selection at i = 1) as a lower bound. All results with $i > 1$ employ neighbor fetching.

\paragraph{Accuracy impact.} We evaluate the accuracy-efficiency tradeoff on RULER at 64K context length. As shown in Table~\ref{tab:prediction_interval}, increasing the prediction interval from every step ($i{=}1$) to every 16 steps ($i{=}16$) reduces average accuracy by only 1.1\% (89.9\% $\to$ 88.8\%), while reducing predictor compute by $16{\times}$. Even at $i{=}8$, accuracy (88.5\%) remains competitive with xKV (88.5\%) from Table~\ref{tab:ruler_st}.

\paragraph{On-GPU efficiency.} Figure~\ref{fig:interval_latency} shows the on-GPU latency impact. The baseline TokenButler ($i{=}1$) incurs predictor overhead comparable to Dense Attention, but with prediction interval this overhead is amortized. At 128K context, $i{=}16$ achieves 30.7ms per token-a 1.6$\times$ speedup over Dense (49.1ms). Notably, prediction interval with neighbor fetching achieves $\approx$1.4$\times$ speedup over Oracle ($i{=}1$), which runs prediction at every decode step similar to DeepSeek Sparse Attention~\citep{deepseekv32}.

\paragraph{CPU-offloading efficiency.} In the CPU-offloading regime ($\geq$256K), the bottleneck shifts to KV data transfer from host memory rather than predictor compute. Consequently, TokenButler with $i{=}1$ already approaches Oracle performance, and increasing the prediction interval does not yield significant additional gains. As shown in Figure~\ref{fig:decoding_efficiency_8b}, by drastically reducing the volume of transferred KV data, TokenButler achieves 7.6$\times$ lower latency than Dense Attention at 1M context. At 256K, this translates from ${\approx}$3.2s per token (Dense) to ${\approx}$0.6s (TokenButler), enabling real-time long-context inference on GPU memory-limited systems.

\begin{table}[t]
    \centering
    \resizebox{\columnwidth}{!}{
    \begin{tabular}{lr|rrrrrrrrrr|r}
    \toprule
    Config & Pred.\ Freq. & N-S1 & N-S2 & N-MK1 & N-MK2 & N-MQ & N-MV & QA-1 & QA-2 & VT & FWE & Avg. \\
    \midrule
    $i{=}1$ & every step & 100.00 & 100.00 & 100.00 & 96.88 & 98.96 & 94.27 & 83.33 & 57.29 & 91.46 & 77.08 & 89.93 \\
    $i{=}2$ & every 2 & 100.00 & 100.00 & 100.00 & 94.79 & 99.22 & 96.88 & 81.25 & 59.38 & 92.50 & 73.96 & 89.80 \\
    $i{=}4$ & every 4 & 100.00 & 100.00 & 100.00 & 94.79 & 97.14 & 93.75 & 81.25 & 57.29 & 92.71 & 72.57 & 88.95 \\
    $i{=}8$ & every 8 & 100.00 & 100.00 & 100.00 & 94.79 & 89.32 & 94.53 & 81.25 & 57.29 & 92.29 & 75.69 & 88.52 \\
    $i{=}16$ & every 16 & 100.00 & 100.00 & 100.00 & 94.79 & 92.19 & 93.49 & 82.29 & 56.25 & 93.33 & 75.69 & 88.80 \\
    \bottomrule
    \end{tabular}
    }
    \caption{RULER evaluation on Llama-3.1-8B-Instruct at 64K context length with prediction interval and neighbor fetching. $i{=}N$ runs the predictor every $N$ steps with neighbor fetching enabled (sparse budget $8K$). Accuracy is maintained within 1.1\% of the per-step baseline ($i{=}1$) even at $16\times$ predictor amortization.}
    \label{tab:prediction_interval}
\end{table}

\section{Conclusion}
We present TokenButler, a lightweight predictor that estimates token importance at fine granularity with minimal overhead. Our co-reference experiments demonstrate that eviction and page-based strategies risk losing critical tokens, whereas TokenButler preserves them with near-oracle accuracy. On long-context benchmarks (RULER, LongBench), TokenButler matches or outperforms prior methods while providing significant latency reductions, especially under CPU offloading. With prediction interval and neighbor fetching, the predictor cost is further amortized while keeping accuracy within 1.1\%, bringing up to 1.6$\times$ speedup over Dense on the GPU and up to 7.6$\times$ speedup in the CPU offloading case. Overall, TokenButler demonstrates that query-aware, fine-grained token selection can replace coarse heuristics with a learned, efficient alternative.

\newpage
\bibliography{colm2026_conference}
\bibliographystyle{colm2026_conference}

\newpage
\appendix

\section{Training Details}
\label{app:training_details}

We keep the TokenButler predictor architecture fixed across backbones: we place a \emph{producer} every \(G=4\) transformer layers (i.e., \texttt{producer\_frequency=4}), predict low-dimensional importance queries with interaction dimension \(d' = 16\), and use a two-layer MLP with hidden size \(512\); the base LLM is frozen and we train only TokenButler parameters with \(\texttt{lr}=10^{-3}\). Unless otherwise noted, predictor training uses training sequence length of 1024 tokens: the training data is a concatenation of (i) general web text from C4 (RealNewsLike, 90k examples)~\cite{c4data}, (ii) educational web text from FineWeb-Edu (\texttt{sample-10BT}, 90k examples)~\cite{finewebdata}, (iii) code from CodeParrot-Clean (90k streamed examples)~\cite{codeparrotdata}, and (iv) long-context QA-style sequences from BABILong~\cite{babilongdata} (contexts \(\{2\mathrm{k},4\mathrm{k},8\mathrm{k},16\mathrm{k}\}\) across tasks \texttt{qa1}-\texttt{qa10}), where each BABILong example is flattened into a single text field (\texttt{input + question + target}) and then tokenized into fixed-length training windows (padding with \texttt{eos} if needed). To make distillation efficient, we compute the cross-entropy loss between teacher and student \emph{masked causal attention distributions} on a subsampled set of query rows: when row-subsampling is enabled we draw most loss rows from the late-context ``tail'' (by default, the final fraction of positions) and always include the final token, reducing auxiliary attention-loss cost from \(\mathcal{O}(L^2)\) to \(\mathcal{O}(RL)\) with \(R\ll L\). We train separate TokenButler predictors for each model discussed in this paper.

\subsection{Training cost}
Training TokenButler does \emph{not} fine-tune the base LLM. On a single A6000 GPU,
the end-to-end predictor training time (including data loading and optimization)
is shown in Table \ref{tab:train_time}.

\begin{table}[ht]
  \centering
  \begin{tabular}{lcccc}
      \toprule
      Model & \multicolumn{2}{c}{TokenButler params} & Training time & Param.\ Percentage \\
      \cmidrule(lr){2-3}
       & (M) & (exact) & (hh:mm) & (\%) \\
      \midrule
    Llama-3.2-1B                 & 8.93  & 8{,}929{,}280  & 04:43 & 0.893 \\
    Llama-3.2-3B                 & 19.77 & 19{,}769{,}344 & 06:45 & 0.659 \\
    Llama-3.1-8B-Instruct        & 29.43 & 29{,}425{,}664 & 09:08 & 0.368 \\
    DeepSeek-R1-Distill-Llama-8B & 29.43 & 29{,}425{,}664 & 09:33 & 0.368 \\
    Qwen2.5-7B-Instruct-1M       & 20.92 & 20{,}923{,}392 & 08:37 & 0.299 \\
    \bottomrule
  \end{tabular}
  \caption{TokenButler predictor training time on a single A6000 GPU. }
  \label{tab:train_time}
\end{table}

\section{Effect of Predictor Attachment Depth}
\label{app:predictor_depth}

\begin{figure*}[hb]
  \centering
  \includegraphics[width=\textwidth]{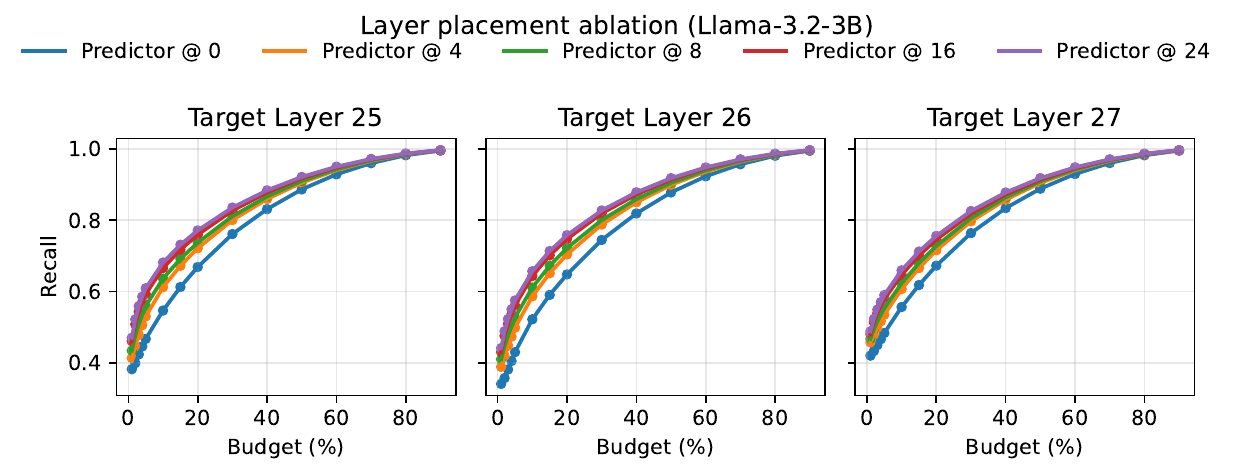}
  \caption{\textbf{Layer placement ablation (Llama-3.2-3B).}
  Recall vs.\ budget for target layers 25–27. Each curve corresponds to a predictor attached at layers \{0, 4, 8, 16, 24\}. Markers denote the measured Recall@\(k\%\) points (e.g., 10/30/50). Later attachment (e.g., predictor @24) consistently yields higher recall across budgets, but leaves fewer layers for sparse execution.}
  \label{fig:predictor_depth}
\end{figure*}

We ablate the layer at which \textsc{TokenButler} consumes hidden states by training five
predictors (each \(\approx\)54.6M parameters) on \texttt{Llama-3.2-3B}, attached at layers
\(\{0, 4, 8, 16, 24\}\). For target layers 25–27, we evaluate recall across a budget sweep
(Recall@\(k\%\)); the resulting curves are shown in Fig.~\ref{fig:predictor_depth}. Plotted
markers correspond to the measured Recall@\(k\%\) values (e.g., 10/30/50), and lines provide a
simple linear interpolation. We find that la ter attachment increases recall across budgets
(predictor @24 is best), but layers \(<k\) must then become dense, reducing the achievable sparsity budget.
In practice, there is a tradeoff such that we (i) attach at a moderate depth to balance recall and sparsity,
or (ii) when memory allows, use multiple lightweight predictors (e.g., every 4 layers) to
approach the accuracy of attaching at later layers, to retain more sparsity.

\section{Predictor Scaling Study}
\label{app:predictor_scaling}

We study how TokenButler's parameter count affects token-importance recovery.
All predictors are trained with the same protocol on \texttt{Llama-3.2-3B} and
evaluated on WikiText2. Table~\ref{tab:scaling_recall50} reports Recall@50\%,
i.e., the fraction of ground-truth high-importance tokens recovered when keeping
the predictor's top-50\% predictions (averaged over heads and sparse layers).

\begin{table}[t]
  \centering
  \begin{tabular}{lcccccc}
    \toprule
    \textbf{Predictor size (M params)} & 3.48 & 5.06 & 12.40 & 39.66 & 144.52 & 287.00 \\
    \midrule
    \textbf{Recall@50\% (\%)} & 67.38 & 70.18 & 71.90 & 73.90 & 79.70 & 81.02 \\
    \bottomrule
  \end{tabular}
  \caption{\textbf{Predictor size scaling (Llama-3.2-3B).} Larger predictors yield higher Recall@50\%.}
  \label{tab:scaling_recall50}
\end{table}

We observe a smooth scaling trend: increasing the predictor
size from 3.48M to 287M improves Recall@50\% by +13.6 points (67.38\% $\to$ 81.02\%),
providing a convenient accuracy/overhead trade-off for different deployment budgets.

\section{Synthetic Co-reference Benchmark}
\label{appx:synthetictask}

To rigorously evaluate token sparsity methods under retrieval-intensive scenarios, we developed a synthetic co-reference benchmark utilizing OpenAI's \texttt{gpt-4o-mini} model. The benchmark consists of 100 unique fictional location names, 100 paired location introductions and tieback questions, 100 philosophical reflections, 100 culinary descriptions, and 100 short math problems. Each data sample is constructed by randomly selecting one location introduction along with its corresponding tieback question, one location name, one philosophical statement, one culinary description, and one math problem. The resulting sequence is structured such that the location is introduced early in the context, followed by distractor content, and concludes with a prelude statement that prompts the retrieval of the original location name.

This modular generation approach allows for the creation of up to $100^{4} = 10^{8}$  unique sequences by combining different components, ensuring extensive diversity. When a specific number of samples are requested, they are dynamically generated by randomly drawing from the respective pools of location introductions, location names, philosophical statements, culinary descriptions, and math problems. This on-the-fly sampling methodology ensures that each test instance presents a distinct retrieval challenge, effectively simulating real-world conversational dynamics where important tokens may reappear unpredictably after various interleaved topics. By designing the benchmark in this manner, we specifically target the capability of token sparsity methods to accurately retain and retrieve critical tokens between substantial contextual noise, thereby providing a robust assessment of their effectiveness in maintaining model performance on co-referential tasks.

\section{Timing Breakdown}
\label{app:timing_breakdown}

\begin{figure}
    \centering
    \includegraphics[width=0.8\linewidth]{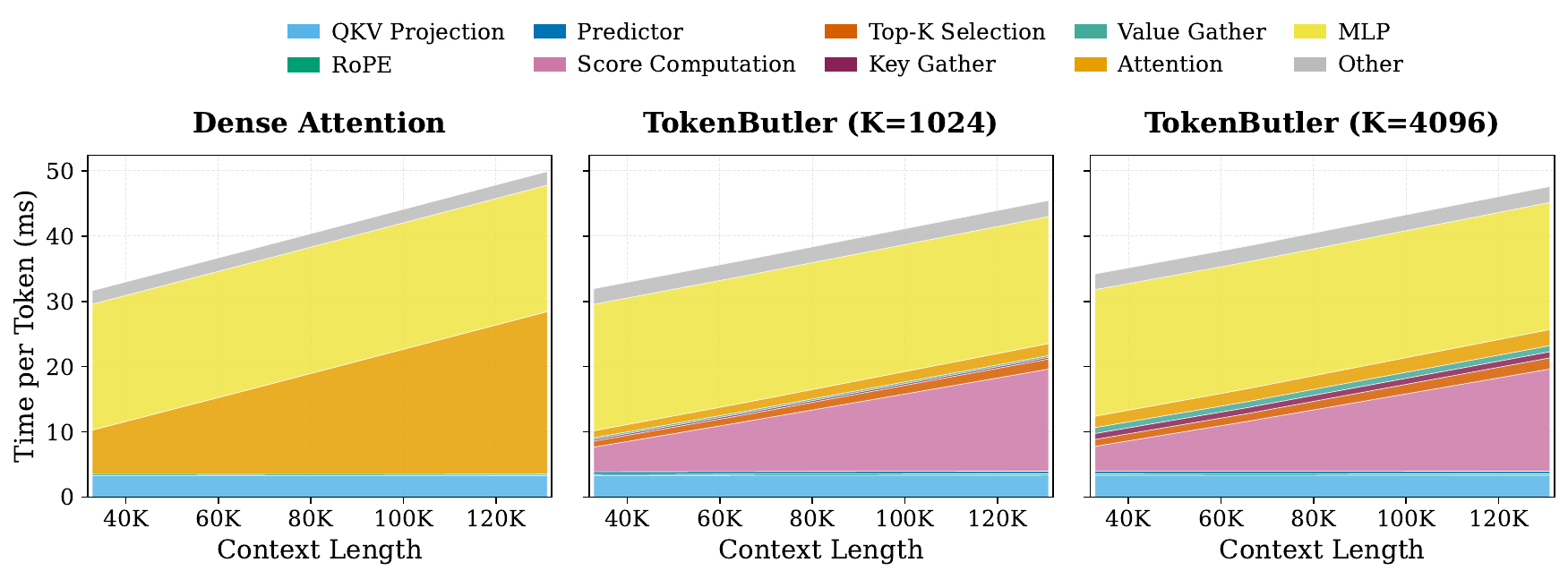}
    \caption{Breakdown of time taken in different operations for \texttt{Llama-3.1-8B-Instruct}.}
    \label{fig:timing_breakdown}
\end{figure}

We analyze the time taken to do different operations in the case of Dense Attention and in case of TokenButler. As shown in Figure~\ref{fig:timing_breakdown}, The Attention Kernel in TokenButler takes almost a constant time as the context length increases while it is the main growth in latency factor in Dense Attention. On the other hand, the most growing operation in TokenButler with context length is the Importance Score Computation because that one has to go through all existing tokens but at a smaller dimension than the original attention. It can be seen also that some of the constant times like gathering of keys and values and the Attention Kernel consume more time when the Sparse Token Budget (K) is higher.

\newpage
\section{Additional Efficiency Evaluation}
\label{app:additional_efficiency_evaluation}

\begin{wrapfigure}[21]{r}{0.45\textwidth}
\vspace{-14pt}   %
\centering
\includegraphics[width=\linewidth]{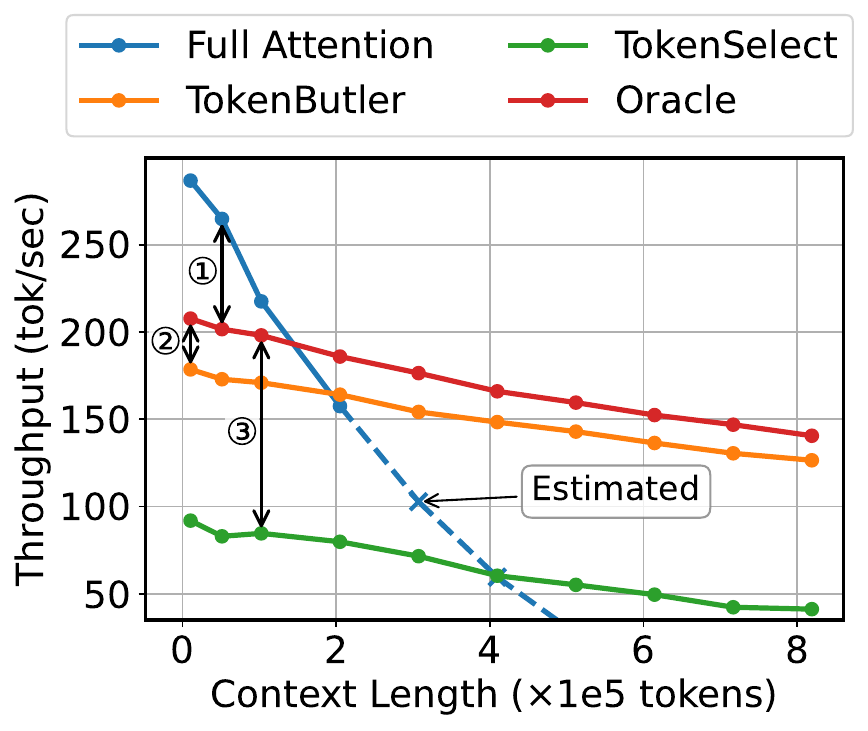}
\caption{Performance of TokenButler vs. Dense Attention and TokenSelect at 1024 token budget on an H100 GPU. \Circled{1}: Sparse Attention Overhead. \Circled{2}: TokenButler Overhead. \Circled{3}: TokenSelect Overhead.}%
\label{fig:performance_1}
\end{wrapfigure}

We provide additional evaluation of the end-to-end performance. We integrate TokenButler with a Llama-3.2 1B model and measure the end-to-end decode throughput under different context lengths in Figure~\ref{fig:performance_1}.
The evaluation utilizes TokenSelect~\citep{wu2024tokenselect} code base where we replace their method by a different version of TokenButler that predicts the importance per token per layer removing the head dimension from the predictions to match the token retrieval method of the system. Full attention throughput drops as the context length increases, eventually giving an error. Token sparsity methods like TokenButler are needed to counter that. TokenButler throughput is close to the oracle performance and TokenButler is more efficient than TokenSelect as our predictor is very lightweight and does not need to do the dot product with the full original embedding dimension $E$ between Q and K.

\end{document}